 \documentclass[preprint,12pt]{elsarticle}

\usepackage{amssymb}
\usepackage{graphicx}
\usepackage{amsmath}
\usepackage{amssymb}
\usepackage{booktabs}
\usepackage{subfig}
\usepackage{xcolor}
\usepackage[pagebackref,breaklinks,colorlinks]{hyperref}
\usepackage{times}
\usepackage{epsfig}
\usepackage{graphicx}
\usepackage{amsmath}
\usepackage{amssymb}
\usepackage{adjustbox}
\usepackage{tabularx}
 \usepackage{verbatim}
\usepackage{bm}
\usepackage{bm}
\usepackage{bbm}
 
 \usepackage{algorithm}

\usepackage{algorithmic}%

\usepackage[capitalize]{cleveref}
\crefname{section}{Sec.}{Secs.}
\Crefname{section}{Section}{Sections}
\Crefname{table}{Table}{Tables}
\crefname{table}{Tab.}{Tabs.}

\journal{Medical Image Analysis}

\begin{document}

\begin{frontmatter}



\title{Unsupervised Model Adaptation for Source-free Segmentation of Medical Images}


\author{Serban Stan and Mohammad Rostami}

\affiliation{organization={University of Southern California}
}
\begin{abstract}
The recent prevalence of deep neural networks has lead semantic segmentation networks to achieve human-level performance in the medical field when sufficient training data is provided. Such networks however fail to generalize when tasked with predicting semantic maps for out-of-distribution images, requiring model re-training on the new distributions. This expensive process necessitates expert knowledge in order to generate training labels. Distribution shifts can arise naturally in the medical field via the choice of imaging device, i.e. MRI or CT scanners. To combat the need for labeling images in a \textit{target domain} after a model is successfully trained in a fully annotated \textit{source domain} with a different data distribution, \textit{unsupervised domain adaptation} (UDA) can be used. Most UDA approaches ensure target generalization by creating a shared source/target latent feature space. This allows a source trained classifier to maintain performance on the target domain. However most UDA approaches require joint source and target data access, which may create privacy leaks with respect to patient information. We propose an UDA algorithm for medical image segmentation that does not require access to source data during adaptation, and is thus capable in maintaining patient data privacy. We rely on an approximation of the source latent features at adaptation time, and create a joint source/target embedding space by minimizing a distributional distance metric based on optimal transport. We demonstrate our approach is competitive to recent UDA medical segmentation works even with the added privacy requisite.\footnote{Early partial results of this work has been presented in 2022 British Machine Vision Conference \cite{stan2021domain}.} 
\end{abstract}






\end{frontmatter}

\section{Introduction}

Semantic segmentation is an area of Computer Vision dedicated to identifying and labeling different parts of an image \cite{GARCIAGARCIA201841}. Segmentation datasets often contain multi-channel images which need to be assigned pixel level labels out of several available semantic categories. In the presence of sufficient annotated images, a classification model can be trained to generalize the task of labeling. However, given the complexity of pixel-level labeling, deep neural networks have shown to be particularly effective at this task, especially convolutional neural networks (CNNs) \cite{lecun1995convolutional}. 

CNNs are designed to exploit image structure via the use of convolution filters, and have been successfully applied to many vision tasks, such as object tracking across scenes ~\cite{yilmaz2006survey,bertinetto2016fully,Zhu_2018_ECCV}, street image segmentation and autonomous vehicle control ~\cite{Kim_2017_ICCV,Hecker_2018_ECCV,stan2021unsupervised}, and in the medical field, for tissue analysis and diagnosis ~\cite{ker8241753deep,shen2017deep,AYACHE2017xxiii,KAZEMINIA2020101938}.   CNNs have been shown to have properties  similar to the visual system~\cite{morgenstern2014properties} and offer human or super-human performances on vision tasks, yet this performance is conditioned on the presence of large amounts of annotated data. Data annotation is a laborious manual task which is a major challenge when using deep learning~\cite{rostami2018crowdsourcing}.  The difficulty of training such networks stems from the large number of parameters being used. Empirically deeper neural networks have performed better than more shallow counterparts at the cost of increased training difficulty \cite{he2016deep}. This reliance on training data has made CNNs sensitive to changes in the input image distribution. Techniques such as dropout \cite{baldi2013understanding} or image augmentations \cite{shorten2019survey} have been designed to counter over-fitting, however the scope of such approaches is limited to scenarios when the application domain, i.e. target domain, of the network is the same as the training domain, i.e. source domain. When the segmentation tasks remains the same but the source and target domains share a different input distribution, e.g. computer generated street images versus real world street images, CNNs experience a considerable performance decay. 

Such differences between source and target distributions, i.e. domain shift, can be overcome naively by gaining access to more labeled data. While for image classification tasks this can be an acceptable outcome, in semantic segmentation labeled images require pixel-level annotations. This makes the labeling process expensive and time consuming \cite{estimating2011liu}. Additionally, labeling medical image data requires the oversight of trained professionals, which further increases the penalty of this process. Unsupervised domain adaptation (UDA) is an area of AI which studies the problem of model generalization on unseen input distributions, and attempts to maintain performance without the need for relabeling samples \cite{ghifary2016deep,venkateswara2017deep,saito2018maximum,zhang2018collaborative,rostami2021lifelong,wu2023unsupervised}. To achieve this objective UDA approaches attempt to produce a shared feature space of source and target embeddings. If this is realised, then a classifier trained on the latent features produces from the source domain will be able to generalize on the unannotated target data. Creating a domain invariant feature space has been explored using adversarial learning \cite{hoffman2018cycada,dou2019pnp,tzeng2017adversarial,bousmalis2017unsupervised,jian2023unsupervised}, where a source and target feature extractor are trained with a GAN discriminator loss \cite{goodfellow2020generative}. Direct distribution minimization between the source and target latent features has also been employed to produce a shared source/target feature space \cite{chen2019progressive,sun2017correlation,lee2019sliced,rost2021unsupervised}. 

UDA benefits from having access to both source and target domains simultaneously \cite{choi2019self,sankaranarayanan2018generate,gabourie2019learning}. This assumption however cannot always be met. In healthcare applications, sharing data is often difficult due to privacy regulations. To maintain the benefits from UDA, source-free adaptation has been developed to bypass the need for direct access to a source domain at adaptation time. While source-free UDA has been previously explored for image classification \cite{ganin2015unsupervised,long2016unsupervised,kang2019contrastive,rostami2023overcoming} and street semantic segmentation \cite{van2021unsupervised,zou2018unsupervised}, there are few works addressing this same problem for medical images analysis \cite{bateson2022source}. Medical images are produced by electromagnetic devices such as MRI or CT scanners, which directly impact the input distribution of the image data. Additionally, compared to natural image semantic segmentation, large portions of medical images are unlabeled. This makes directly applying segmentation algorithms designed for street semantic segmentation unsuitable, as we later show in our experimental section. 

We propose a new source-free semantic segmentation algorithm for medical images that relies on the idea of distributional distance minimization. After source training, we learn a sampling distribution to approximate the source latent embeddings. During adaptation we lose access to the source data, and utilize this distribution as a surrogate. We perform direct distributional alignment between target embeddings and the sampling distribution by using an optimal transport based metric. We provide empirical evidence that our algorithm is competitive with state of the art (SOTA) medical UDA works by exploring domain adaptation problems on a cardiac and organ dataset. We additionally provide theoretical justification for the performance of our approach by developing an upper bound for expected target error using optimal transport theory.  

\section{Related Work}

Deep neural networks designed for semantic segmentation usually follow a common structure of feature encoder, up-sampling decoder followed by a classification head \cite{chen2017deeplab}. The feature encoder component summarizes the input data distribution to a latent feature distribution with reduced dimensionality. The effect of data summarization allows for image-level noise to be removed from the prediction process. An up-sampling decoder then maps the latent embedding space to an embedding space the same size of the input space. This step is necessary due to the fact that segmentation problems require pixel-level labels. Shortcut links between encoder and decoder layers can offer an improved quality in the up-sampling process \cite{10.1007/978-3-319-24574-4_28}. A classification network then learns a decision function on the pixel-level logits. Domain adaptation approaches for semantic segmentation rely on creating a shared embedding space between the source and target. This has been explored at several levels of abstraction in the latent feature space \cite{saito2018maximum,pan2019transferrable}. Creating a domain-agnostic feature distribution is the primary focus of domain adaptation approaches. 

A large number of UDA approaches rely on ideas from adversarial learning to reduce domain gap between source and target embeddings. One solution for this problem comes in the form of reducing the domain gap at the level of the input space. Style-transfer networks \cite{jing2019neural} have been developed to visually translate images to different domains, e.g. turning a street photo into a painting in a specific artist's style \cite{Gatys_2016_CVPR}. In the case of UDA, learning a style transfer network between the source domain and target domain can offer an extension of the source dataset in the style of the target domain, which allows the source classifier to directly generalize on target samples \cite{hoffman2018cycada}. A second set of approaches directly use a discriminator loss to ensure the source and target samples are embedded in a shared space \cite{saito2018maximum}. A discriminator network attempts to verify whether pairs of embeddings are from the same domain or from the source and target, and an encoder network is trained to increase the difficulty of this decision task. While adversarially trained networks have proven their effectiveness in UDA \cite{Zhang_2018_CVPR,10.1007/978-3-319-59050-9_47}, training such networks requires access to large amounts of data and careful hyper-parameter fine-tuning. This can pose significant downsides in the case of source-free UDA, where joint access to the source and target domains is not permitted.  

Relating AI problem in latent embedding spaces can  be done by directly realizing distributional relations to transfer information~\cite{rostami2021transfer}. This idea has been explored in UDA by minimizing the distance between the source and the target distributions  \cite{drossos2019unsupervised,LE2019249,liu2020pdam,rostami2021cognitively}, and aims to produce a domain-agnostic embedding space by ensuring the source and target latent distributions are indistinguishable with respect to some chosen distributional distance. Previous choices for this distance have been the mean statistic \cite{yan2017mind}, KL-divergence \cite{toth2016adaptation} or optimal transport \cite{lee2019sliced,stan2021secure}. Arriving at the appropriate distributional distance metric is still an active area of research, however recently Wasserstein Distance (WD) has shown desirable properties in high dimensional optimization tasks \cite{frogner2015learning, LE2019249, kolouri2019generalized}. However, the WD does not provide a closed-form solution in the general case. This requires solving a non-linear optimization problem to correctly compute the WD. Our proposed approach is also based on minimizing distributional discrepancy between the source and target domains. In our algorithm, we utilized the Sliced Wasserstein Distance (SWD), an approximation of the WD that allows for end-to-end gradient-based optimization and has been shown to be effective in UDA settings~\cite{lee2019sliced,rostami2019deep}. 

Most UDA methods are designed with the premise of joint source/target access. This has been true for UDA methods in the medical field as well \cite{huo2018adversarial,zhang2018task,chen2018semantic,kamnitsas2017unsupervised}, where restrictions in data privacy can lead to different databases being accessible sequentially. This problem has been previously explored in source-free UDA, where access to the source data is lost following source training. Most segmentation works exploring source-free UDA have been targeting real image segmentation \cite{Kundu_2021_ICCV, liu2021sourcefree}, however the distributional discrepancy between street images and medical images does not allow for direct application of these algorithms to the medical field, as we later demonstrate in our experimental section. In the medical field, distributional distance minimization approaches have been explored for minimizing the KL-divergence between latent feature distributions \cite{BatesonSFDA,bateson2022source}. We compare with these recent works and show our optimal transport based approach offers improved adaptation performance. 

\section{Problem Formulation}

\begin{figure*}[!htb]
    \centering
     \includegraphics[width=.8\textwidth]{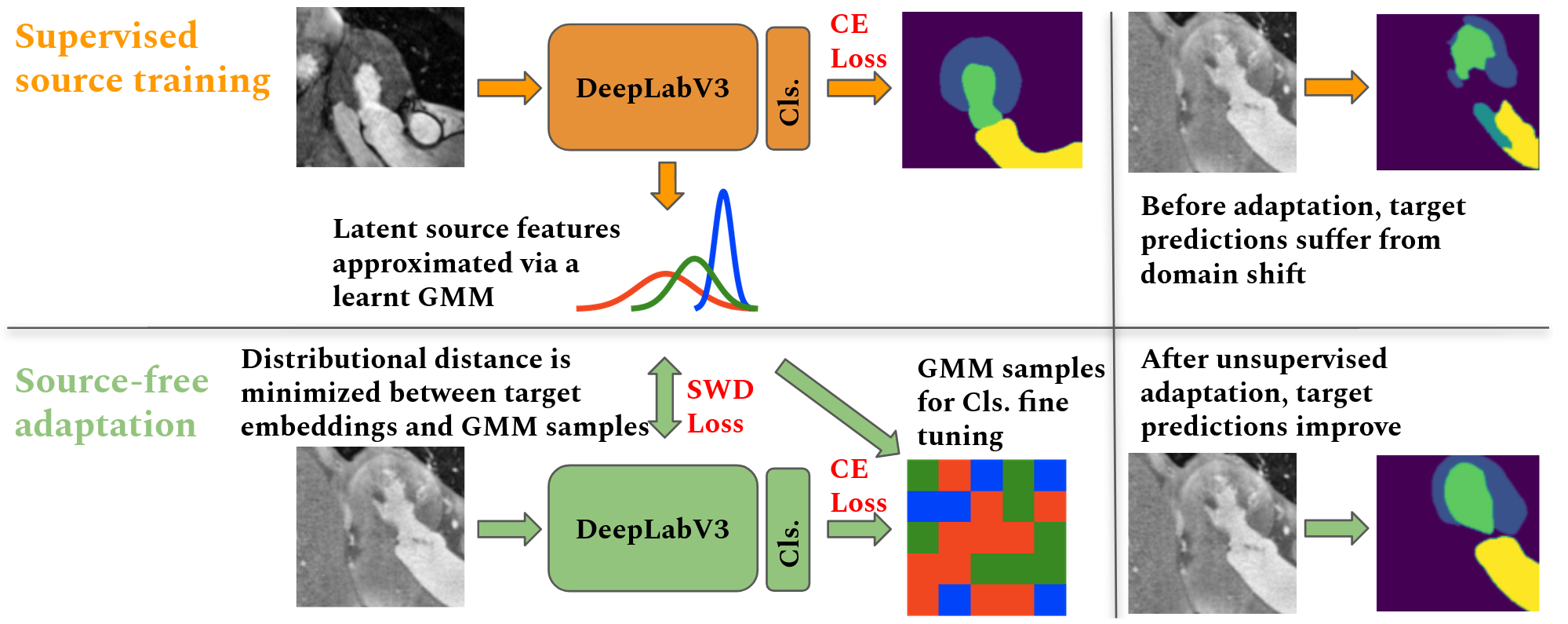}
    \caption{\small Proposed method: We first perform supervised training on source MR images. Using the source embeddings we characterize an internal distribution via a GMM distribution in the latent space. We then perform source free adaptation by matching the embedding of the target CT images to the learnt GMM distribution, and fine tuning of the classifier on GMM samples. Finally, we verify the improved performance that our model gains from model adaptation. }
     \label{figure:problem-formulation}
\end{figure*}

Consider the available inputs on the source and target domains to be slices of 3D organ scans. Following literature \cite{chen2019synergistic}, we assume each image to appear in a three channel format, i.e. three consecutive slices. The pixel labels will correspond to the middle slice. Let the source dataset be $D^S = \{ (x^s_1,y^s_1), (x^s_2, y^s_2) \dots (x^s_N, y^s_N) \}$, and the target dataset be $D^T = \{ (x^t_1,y^t_1), (x^t_2, y^t_2) \dots (x^t_M,y^t_M) \}$, where $(x^s_i, y^s_i)$ are source domain image/label pairs, and  $(x^t_i, y^t_i)$ have similar meaning for the target domain. Also, let $X_S$ be the set of all source images, and $X_T$ be the set of all target images, i.e. $X_S = \{ x^s_1, x^s_2 \dots \}$, $X_T = \{ x^t_1, x^t_2 \dots \}$. We assume each image shares the same input dimension of $W \times H \times 3$, however the probability distributions of the source images $\mathcal{P_S}$ and the target images $\mathcal{P_T}$ differ. The label space between the source and target domains is the same, i.e. each pixel is labeled with one of $K$ semantic classes. 
Under the restrictions of UDA, our task is to learn a segmentation model $\phi$ with learnable parameters $\theta$ using the fully supervised source domain $D_S$ and allow it to generalize on the target domain $D_T$, where only access to unlabeled images $X_T$ is possible. We structure $\phi$ as a semantic segmentation network composed of a feature extractor $f$, an up-sampling decoder $g$ and a probabilistic classifier $h$, such that $\phi = f \circ g \circ h$. The model output of $\phi$ returns pixel-wise class probabilities as $K$-dimensional vectors, where $K$ is the number of semantic classes. 

Initially, the segmentation model learns an appropriate decision function on the labeled source domain. This can be done by minimizing empirical risk with respect to the available source samples, by minimizing a pixel-level cross-entropy loss:

\begin{equation}
    \hat\theta = \arg\min_{\theta} \frac{1}{N} \sum_{i=1}^N \mathcal{L}_{CE} (y^s_i, \phi_\theta(x^s_i))
    \label{eq:source-training}
\end{equation}

where $\mathcal{L}_{CE}$ is $K$ class cross entropy loss:

\begin{equation}
    \mathcal{L}_{CE} (y, \hat{ y}) = -\frac{1}{W H} \sum_{i=1}^W \sum_{j=1}^H \sum_{k=1}^K \mathbbm{1}_{k}(y_{i,j}) \log \hat{y_{i,j,k}}
    \label{eq:ce-loss}
\end{equation}

The notation $\mathbbm{1}_k(x)$ represents the indicator function which returns $1$ if $x$ equals to $k$ and zero otherwise.

After learning $\hat\theta$, our model will be able to generalize on new samples from $\mathcal{P_S}$. Our goal is to minimize the expected risk on the target distribution $\mathcal{P_T}$ without access to target labels. To achieve this, we aim to create a feature space of classifier inputs that is common for the source and target domains. This translates to $f \circ g(X_S) \approx f \circ g(X_T)$. UDA approaches have explored direct alignment between two distributions, by choosing an appropriate metric to be minimized \cite{wu2018_dcan_eccv,zhang2017_curriculum}. In our work, we choose the SWD as this metric, which allows for gradient pass-through for network optimization and has been shown to be an effective metric for distribution alignment. Thus, we will minimize $SWD(f \circ g (X_S), f \circ g (X_T))$.

As our setting is concerned with maintaining the privacy of patient records, we do not assume direct access to the dataset $D_S$ during adaptation. Thus, in order for the above approach to be viable, we require an approximation of the source encodings $f \circ g(X_S)$. We produce this by learning a set of Gaussians to fit the source latent embedding data. Under the assumption of sufficient training samples, a GMM distribution will be able to accurately approximate the underlying data distribution at the decoder output. This allows us to still use the SWD for domain alignment without needing access to the source domain during adaptation. 

\section{Proposed Algorithm}
\label{sec:proposedalgorithm}

We present a visual description of our approach in Figure \ref{figure:problem-formulation}. The first step in our approach is to train the network $\phi$ on the available source data. Once training is fully complete, we can approximate the distribution at the decoder output, i.e. $f \circ g$ as a GMM. After this step, we no longer require direct access to the source distribution. Once adaptation commences, we initialize the network with source weights, and minimize the SWD distance between samples from the GMM and target embeddings at the decoder output. Given the GMM distribution differs from the source embeddings, we additionally fine tune the classifier on GMM samples. Below, we provide in-depth descriptions of the various components of this approach.

After source training, we model the source feature embeddings via a GMM distribution. Let the source embedding distribution be $\mathcal{P}_Z$. As the segmentation problem we consider has $K$ semantic classes, we expect our embedding space to be separable into $K$ clusters, each feature cluster corresponding to one of the semantic classes. In our GMM approximation we will learn each of the clusters independently, via $t$ Gaussians. This can be achieved in practice as the source domain labels are available at GMM learning time. The approximation of $\mathcal{P}_Z$ is computed as follows:

\begin{equation*}
    \mathcal{P_Z}(z) = \sum_{i=1}^{t \cdot K} \alpha_i p_i(z)
        = \sum_{i=1}^{t \cdot K} \alpha_i \mathcal{N} (z | \mu_i, \Sigma_i),
\end{equation*}

$\alpha_i, \mu_i, \Sigma_i$ represent the coefficients, means and covariance matrices of the learnt Gaussians. We learn the GMM distribution via the Expectation Maximization algorithm, however we can improve the learning process by making use of the source labels. For each semantic class $k \in \{1 \dots K\}$ we identify which latent embeddings correspond to label $k$. This in turn permits us to learn the Gaussians in an unsupervised way only with samples that are already aligned to a specific class. We thus avoid the possibility of samples from different classes negatively impacting the learning process, which may result in a Gaussian corresponding to several semantic classes. 

We can further tune the GMM learning process by establishing a confidence threshold for considered data samples. Each of the data points used to learn the GMMs are decoder outputs, thus the classifier $h$ can assign them a confidence score. If for some sample this confidence score is low, then the sample will be closer to the decision boundary of the classifier, whereas if the confidence score is high, the sample will be a good representation of the feature cluster. We introduce a confidence parameter $\rho$ to enforce this observation, which allows us to avoid samples close to the decision boundary in the learning process. Thus, for each class $k$, the $t$ Gaussians in our GMM will be learned from the set of embeddings $S_{\rho,k} = \{ f \circ g (x_i)_{p,q} | \phi (x_i)_{p,q} > \rho, \arg\max \phi (x_i)_{p,q} = (y_i)_{p,q}\}$, where $(p,q)$ are pairs of pixel positions in a $W \times H$ image. 
Note in the above formulation one Gaussian per semantic class is the minimum requirement in order to learn the GMM. This corresponds to $t=1$. However, our algorithm is not limited in this regard, and we can choose a larger number of Gaussians for class representation. A larger value of $t$ will improve the approximation of the feature embeddings, however will come with a downside of larger training time. However, the training process is dominated by the optimization of the cross-entropy loss in Eq. \ref{eq:source-training}. We choose $t=3$ in our reporting, and observe a beneficial impact on the two main metrics we consider. 

Once the GMM parameters are learned we no longer require access to the source domain. For adaptation, we minimize the distributional distance between the target distribution and the GMM distribution. In practice, we will optimize the SWD at the batch level between target samples and samples from the GMM. To this end, we create a dataset $D_P = (Z_P, Y_P)$ of GMM samples. The SWD loss acts as an approximation of the WD for high dimensional distributions. The SWD distance averages over $V$ evaluations of the WD for 1-dimensional projections of the two distributions. As the WD for one dimensional distributions has a closed form solution, minimizing the SWD allows for end-to-end gradient based optimization of the $\phi$ network. The general formula for the SWD is as follows:

\begin{equation}
    \mathcal{L}_{SWD}(P,Q) = \frac{1}{V} \sum_{i=1}^V WD(\langle \gamma_i, P \rangle, \langle \gamma_i, Q \rangle)
    \label{eq:swd}
\end{equation}

where $\gamma_i$ is a random projection direction. 

The final adaptation loss is thus composed of two terms. The first is the cross entropy loss with respect to samples from $D_P$ and their labels $Y_P$ for fine-tuning the classifier. The second term is the SWD loss presented in Eq. \ref{eq:swd} between samples from $D_P$ and target image embeddings, $f \circ g (X_T)$. We can express the adaptation loss, $\mathcal{L}_{adapt}$, as follows:

\begin{equation}
    \begin{split}
        \mathcal{L}_{adapt} = \mathcal{L}_{CE}(Z_P, Y_P) + \lambda \mathcal{L}_{SWD}(Z_P, f \circ g(X_T))
    \end{split}
    \label{eq:loss-adapt}
\end{equation}

for some regularizer $\lambda$. The pseudocode for our approach, called Source Free semantic Segmentation (SFS), is present in Alg. ~\ref{SSUDAalgorithmSS}.

\begin{algorithm}[t]
\caption{$\mathrm{SFS}\left (\lambda , \rho \right)$\label{SSUDAalgorithmSS}} 
 {
    \begin{algorithmic}[1]
    \STATE \textbf{Initial Training}: 
    \STATE \hspace{2mm}\textbf{Input:} Source domain dataset $\bm{D_S}=(\bm{X_S},  \bm{Y_S})$,
    \STATE \hspace{4mm}\textbf{Training on Source Domain:}
    \STATE \hspace{4mm} $\hat{ \theta}=\arg\min_{\theta}\sum_i \mathcal{L}(f_{\theta}(\bm{x}_i^s),\bm{y}_i^s)$
    \STATE \hspace{2mm}  \textbf{Internal Distribution Estimation:}
    \STATE \hspace{4mm} Set $\rho=.97$, compute $\mathcal{S}_{\rho, k}$ and estimate $\hat\alpha_j, \hat\mu_j,$ and $\hat\Sigma_j$ class conditionally
    \STATE \textbf{Model Adaptation}: 
    \STATE \hspace{2mm} \textbf{Input:} Target dataset $\bm{D_T}=(\bm{X_T})$
    \STATE \hspace{2mm} \textbf{Pseudo-Dataset Generation:} 
    \STATE \hspace{4mm} $\bm{D_P}=(\bm{Z_P},\bm{Y_P})=$
    $([\bm{z}_1^p,\ldots,\bm{z}_N^p],[\bm{y}_1^p,\ldots,\bm{y}_N^p])$, where:
    $\bm{z}_i^p\sim \mathcal{P}_Z(\bm{z}), 1\le i\le N_p$
    \FOR{$itr = 1,\ldots, ITR$ }
    \STATE Draw random batches from $\bm{D_T}$ and $\bm{D_P}$
    \STATE Update the model by solving Eq.~\eqref{eq:loss-adapt}
    \ENDFOR
    \end{algorithmic}}
\end{algorithm} 

\section{Theoretical Analysis}
\label{sec:theoreticalanalysis}

We prove that Algorithm~\ref{SSUDAalgorithmSS} is effective because an upper-bound on the expected error for the target domain is minimized as a result of domain alignment.

We analyze the problem in a standard PAC-learning setting. Consider that the set of  classifier sub-networks $\mathcal{H} = \{\phi_{\bm{w}}(\cdot)|\phi_{\bm{w}}(\cdot):\mathcal{Z}\rightarrow \mathbb{R}^k, \bm{w}\in \mathbb{R}^W\}$ to be our hypothesis space. Let $e_{\mathcal{S}}$ and  $e_{\mathcal{T}}$ denote the   expected error of the optimal   model that belongs to this space on the source and target domains, respectively.  Let $\phi_{\bm{w}^*}$ to be the model  which minimizes the combined source and target expected error $e_{\mathcal{C}}(\bm{w}^*)$, defined as: $\bm{w}^*= \arg\min_{\bm{w}} e_{\mathcal{C}}(\bm{w})=\arg\min_{\bm{w}}\{ e_{\mathcal{S}}+  e_{\mathcal{T}}\}$. This model is the best model within the hypothesis space in terms generalizability in both domains. Additionally, consider that  $\hat{\mu}_{\mathcal{S}}=\frac{1}{N}\sum_{n=1}^N f \circ g(\bm{x}_n^s)$ and $\hat{\mu}_{\mathcal{T}}=\frac{1}{M}\sum_{m=1}^M f \circ g(\bm{x}_m^t)$ are the empirical source distribution and the empirical target distribution in the embedding space, both built from the available data points. Let $\hat{\mu}_{\mathcal{P}}=\frac{1}{N_p}\sum_{q=1}^{N_p} {z}_q$ denote the empirical internal distribution, built from the generated pseudo-dataset.  

\textbf{Theorem 1}: Consider that we generate a pseudo-dataset  using the internal distribution and preform UDA according to algorithm~\ref{SSUDAalgorithmSS}, then: 
\begin{equation}
\small
\begin{split}
e_{\mathcal{T}}\le & e_{\mathcal{S}} +W(\hat{\mu}_{\mathcal{S}},\hat{\mu}_{\mathcal{P}})+W(\hat{\mu}_{\mathcal{T}},\hat{\mu}_{\mathcal{P}})+(1-\rho)+e_{\mathcal{C'}}(\bm{w}^*)\\&+\sqrt{\big(2\log(\frac{1}{\xi})/\zeta\big)}\big(\sqrt{\frac{1}{N}}+\sqrt{\frac{1}{M }}+2\sqrt{\frac{1}{N_p }}\big),
\end{split}
\label{theroemforPLnipssemantic}
\end{equation}    
where $W(\cdot,\cdot)$ denotes the WD distance and   $\xi$ is a constant, dependent on the loss function $\mathcal{L}(\cdot)$.

According to Theorem~1,  Algorithm~\ref{SSUDAalgorithmSS} minimizes the upper bound expressed in Eq.~\eqref{theroemforPLnipssemantic} for the target domain expected risk. The source expected risk is minimized when we train the model on the source domain. The second term in Eq.~\eqref{theroemforPLnipssemantic} is minimized when the GMM is fitted on the source domain distribution.  The third term in the Eq.~\eqref{theroemforPLnipssemantic} upperbound is minimized  because it is the second term in Eq.~\eqref{eq:loss-adapt}. The fourth term is small if we let $\rho\approx 1$. 
  The term $e_{C'}(\bm{w}^*)$ will be small if the domains are related to the extent that a joint-trained  model can generalize well on both domains, e.g., there should not be label mismatch between similar classes across the two domains.   The last term in Eq.~\eqref{theroemforPLnipssemantic} is negligible  if  the training datasets are large enough.

 \textbf{Proof:} We use a result   by Redko et al.~\cite{redko2017theoretical} which is developed for domain adaptation based on joint training.

\textbf{Theorem~2 (Redko et al.~\cite{redko2017theoretical})}: Consider that a model is trained on the source domain, then for any $d'>d$ and $\zeta<\sqrt{2}$, there exists a constant number $N_0$ depending on $d'$ such that for any  $\xi>0$ and $\min(N,M)\ge \max (\xi^{-(d'+2),1})$ with probability at least $1-\xi$, the following holds:
\begin{equation}
\begin{split}
e_{\mathcal{T}}\le & e_{\mathcal{S}} +W(\hat{\mu}_{\mathcal{T}},\hat{\mu}_{\mathcal{S}})+e_{\mathcal{C}}(\bm{w}^*)+ \\& \sqrt{\big(2\log(\frac{1}{\xi})/\zeta\big)}\big(\sqrt{\frac{1}{N}}+\sqrt{\frac{1}{M}}\big).
\end{split}
\label{segmenteq:theroemfromcourty}
\end{equation}    
 
Theorem~2 relates the performance of  a source-trained  model on a target domain through an upperbound which depends on the distance between the source and the target domain distributions in terms WD distance. We use Theorem~2 to deduce Theorem~1 in the paper.  Redko et al.~\cite{redko2017theoretical} provide their analysis  for the case of binary classifier but their analysis can be extended to a multi-class scenario.

 Since we use the   parameter $\rho$ to estimate the internal distribution, the probability of  predicting incorrect  labels for the drawn   pseudo-data points   is $1-\rho$.  We  can  define the following difference:
\begin{equation}
\begin{split}
  |\mathcal{L}(h_{\bm{w}_0}(\bm{z}^p_i),\bm{y}^p_i)- \mathcal{L}(h_{\bm{w}_0}(\bm{z}^p_i),\hat{\bm{y}}_i^{p})|= \begin{cases}
    0, & \text{if $\bm{y}^p_i=\hat{\bm{y}}_i^{p}$}.\\
    1, & \text{otherwise}.
  \end{cases}
\end{split}
\label{segmenteq:theroemforPLproof}
\end{equation}    
 We can apply Jensen's inequality following taking the expectation  with respect to the target domain distribution in the embedding space, i.e., $f \circ g(\mathcal{P_T})$, on both sides of Eq.~\eqref{segmenteq:theroemforPLproof} and conclude:
\begin{equation}
\begin{split}
&|e_{\mathcal{P}}-e_{\mathcal{T}}|\le\\&\mathbb{E}\big(|\mathcal{L}(h_{\bm{w}_0}(\bm{z}^p_i),\bm{y}^p_i)- \mathcal{L}(h_{\bm{w}_0}(\bm{z}^p_i),\hat{\bm{y}}_i^{p})|\big)\le   
(1-\rho).
\end{split}
\label{segmenteq:theroemforPLproofexpectation}
\end{equation}    
Now we use Eq.~\eqref{segmenteq:theroemforPLproofexpectation} to deduce:
\begin{equation}
\begin{split}
&e_{\mathcal{S}}+e_{\mathcal{T}}=e_{\mathcal{S}}+e_{\mathcal{T}}+e_{\mathcal{P}}-e_{\mathcal{P}}\le  
e_{\mathcal{S}}+e_{\mathcal{P}}+|e_{\mathcal{T}}-e_{\mathcal{P}}|\le\\&  
e_{\mathcal{S}}+e_{\mathcal{P}}+(1-\rho).
\end{split}
\label{segmenteq:theroemforPLprooftrangleinq}
\end{equation}    
 Taking infimum on both sides of Eq.~\eqref{segmenteq:theroemforPLprooftrangleinq} and employing the definition of the joint optimal model yields:
\begin{equation}
\begin{split}
e_C(\bm{w}^*)\le e_{\mathcal{C'}}(\bm{w})+(1-\rho).
\end{split}
\label{segmenteq:theroemforPLprooftartplerror}
\end{equation}    
Now we consider
Theorem~2  by Redko et al.~\cite{redko2017theoretical} for the source and target domains in our problem and  merge Eq.~\eqref{segmenteq:theroemforPLprooftartplerror}  on Eq.\eqref{segmenteq:theroemfromcourty} to conclude:
\begin{equation}
\begin{split}
e_{\mathcal{T}}\le & e_{\mathcal{S}} +W(\hat{\mu}_{\mathcal{T}},\hat{\mu}_{\mathcal{S}})+e_{\mathcal{C'}}(\bm{w}^*)+ (1-\rho) \\&+ \sqrt{\big(2\log(\frac{1}{\xi})/\zeta\big)}\big(\sqrt{\frac{1}{N}}+\sqrt{\frac{1}{M}}\big).
\end{split}
\label{segmenteq:theroemfromcourty1}
\end{equation}    
In Eq.~\eqref{segmenteq:theroemfromcourty1}, $e_{\mathcal{C'}}$ denotes the joint optimal model true error for the source domain and the pseudo-dataset as the second domain.

Now we apply the triangular inequality twice in Eq.~\eqref{segmenteq:theroemfromcourty1} to  deduce:
\begin{equation}
\begin{split}
& W(\hat{\mu}_{\mathcal{T}},\hat{\mu}_{\mathcal{S}})\le  W(\hat{\mu}_{\mathcal{T}},\mu_{\mathcal{P}})+W(\hat{\mu}_{\mathcal{S}},\mu_{\mathcal{P}})  \le\\& W(\hat{\mu}_{\mathcal{T}},\hat{\mu}_{\mathcal{P}})+W(\hat{\mu}_{\mathcal{S}},\hat{\mu}_{\mathcal{P}})+2W(\hat{\mu}_{\mathcal{P}},\mu_{\mathcal{P}}) .
\end{split}
\label{segmenteq:theroemfromcourty2}
\end{equation}

 We now need Theorem 1.1  by Bolley et al.~\cite{bolley2007quantitative} to  simplify the term $W(\hat{\mu}_{\mathcal{P}},\mu_{\mathcal{P}})$ in Eq.~\eqref{segmenteq:theroemfromcourty2}.

  \textbf{Theorem~3} (Theorem 1.1 by Bolley et al.~\cite{bolley2007quantitative}): consider that 
  
  $p(\cdot) \in\mathcal{P}(\mathcal{Z})$ and $\int_{\mathcal{Z}} \exp{(\alpha \|\bm{x}\|^2_2)}dp(\bm{x})<\infty$ for some $\alpha>0$. Let $\hat{p}(\bm{x})=\frac{1}{N}\sum_i\delta(\bm{x}_i)$ denote the empirical distribution that is built from the samples $\{\bm{x}_i\}_{i=1}^N$ that are drawn i.i.d from $\bm{x}_i\sim p(\bm{x})$. Then for any $d'>d$ and $\xi<\sqrt{2}$, there exists $N_0$ such that for any $\epsilon>0$ and $N\ge N_o\max(1,\epsilon^{-(d'+2)})$, we have:
 \begin{equation}
\begin{split}
P(W(p,\hat{p})>\epsilon)\le \exp(-\frac{-\xi}{2}N\epsilon^2)
\end{split}
\label{segmenteq:mainSuppICML3}
\end{equation}  
 This theorem provides a relation to measure the distance between the estimated empirical distribution and the true distribution when the distance is measured by the WD metric.

 We can use both   Eq.~\eqref{segmenteq:theroemfromcourty2} and Eq.~\eqref{segmenteq:mainSuppICML3} in Eq.~\eqref{segmenteq:theroemfromcourty1} to  conclude Theorem~2 as stated in the paper:
\begin{equation}
\small
\begin{split}
e_{\mathcal{T}}\le & e_{\mathcal{S}} +W(\hat{\mu}_{\mathcal{S}},\hat{\mu}_{\mathcal{P}})+W(\hat{\mu}_{\mathcal{T}},\hat{\mu}_{\mathcal{P}})+(1-\rho)+e_{\mathcal{C'}}(\bm{w}^*)\\&+\sqrt{\big(2\log(\frac{1}{\xi})/\zeta\big)}\big(\sqrt{\frac{1}{N}}+\sqrt{\frac{1}{M }}+2\sqrt{\frac{1}{N_p }}\big),
\end{split}
\label{segmenteq:theroemforPLnips55}
\end{equation}     

 \section{Experimental Validation}
\label{sec:experimentalvalidation}

\subsection{Datasets}

To evaluate our approach we consider two semantic segmentation datasets for internal organ imaging. We describe the two datasets and the adaptation tasks as follows:

\textbf{Multi-Modality Whole Heart Segmentation Dataset (MMWHS)~\cite{ZHUANG201677}:}  the dataset covers 3D scans of heart tissue, and evaluates algorithms on correctly segmenting the scans with respect to four semantic classes: ascending aorta (AA), left ventricle blood cavity (LVC), left atrium blood cavity (LAC), myocardium of the left ventricle (MYO). Elements which do not fall into one of the above four categories are to be ignored in the training process. The $3D$ scans provided are obtained from two types of magnetic imaging devices, $20$ obtained via MRI imaging and $20$ obtained via CT imaging. The adaptation tasks we consider assumes the source domain to be the MRI domain and the target domain to be the CT domain, and the reverse problem, where we adapt from CT to MRI. The MMWHS dataset has been previously considered in the UDA literature for medical image segmentation. We use the data splits prepared and provided by Chen et al.~\cite{chen2020unsupervised}. The data preparation process involves turning each $3D$ scan into a series of $2D$ images. Initially, the values of the pixels are normalized to a standard normal distribution. Then, $3$-channel $2D$ images are obtained by considering three consecutive slices of the $3D$ scan. The semantic labels correspond to the middle slice. Additional data augmentation techniques such as rotations or cropping are used. 



\textbf{CHAOS MR and Multi-Atlas Labeling Beyond the Cranial Vault:} the segmentation problem pertains to the segmentation of abdominal organs into four semantic classes: Liver (L), Right Kidney (RK), Left Kidney (LK) and Spleen (S). We use two publicly available organ segmentation datasets, the 2019 CHAOS MR dataset \cite{CHAOSdata2019} used in the 2019 CHAOS Grad Challenge, and the Multi-Atlas Labeling Beyond the Cranial Vault MICCAI 2015 Challenge dataset \cite{landman2015multiatlas}. The CHAOS MR dataset consists of MRI scans, while the second dataset consists of CT scans. The data preparation process is done similarly to the MMWHS heart segmentation case. The CT images were clipped outside of the range $[-125,275]$ \cite{zhou2019prioraware}, and then normalized to zero mean and unit variance at the pixel level. From the $3D$ scans, $2D$ images were produced by following the slicing process used in the MMWHS case. Four types of augmentation were employed: rotations, value negation, cropping and adding noise. We again consider two adaptation tasks, from the MRI domain to the CT domain, and from the CT domain to the MRI domain.

\subsection{Evaluation Methodology}

We use two metrics for evaluating the quality of adaptation methods for medical image segmentation: Dice coefficient and Average Symmetric Surface Distance (ASSD). The Dice coefficient is an indicator of the quality of segmentation in terms of area, and is used as a primary metric in medical segmentation works \cite{chen2019synergistic,chen2020unsupervised,zhou2019prioraware}. A large Dice score will signify that on average, there is high overlap between model predictions and actual labels. However, the detail of region borders can suffer even in the presence of a large Dice score. As the quality of representations can be especially important in the medical field, the ASSD score is used to measure the fidelity of the predicted semantic map borders. A low value of ASSD signifies that the produced semantic maps are similar in border representation to the actual areas of organ split. 


For both adaptation problems, we compare our algorithm with recent UDA approaches for semantic segmentation. We consider PnP-AdaNet \cite{dou2019pnp}, SynSeg-Net \cite{Huo_2019}, AdaOutput \cite{Tsai_adaptseg_2018}, CycleGAN \cite{zhu2020unpaired}, CyCADA \cite{hoffman2018cycada}, SIFA \cite{chen2020unsupervised}, ARL-GAN \cite{chen2020anatomy}, DSFN \cite{ijcai2020-455}, SASAN \cite{tomar2021self}, DSAN \cite{han2021deep}. The above methods are based on adversarial learning and do not work in a source-free regime, giving them an advantage over our approach. This is due to the fact that these methods benefit of having full access to the source domain during adaptation, and do not need to employ an approximation of the source or some other procedure. We also compare our approach to two source-free medical image segmentation algorithms that designed using the MMWHS datasets: AdaEnt~\cite{BatesonSFDA} and AdaMI~\cite{bateson2022source}. Finally, we support the claim that street image segmentation approaches are not suitable to be directly applied to medical images. We consider GenAdapt \cite{Kundu_2021_ICCV}, a recent street semantic segmentation work with SOTA performance that is designed for street image segmentation, however which does not generalize well to the medical field. We show that our algorithm, even under the added constraint of not having access to the source domain during adaptation, is able to compete and outperform the above mentioned approaches, demonstrating the effectiveness of our adaptation procedure.


\subsection{Experimental Setup}

We follow common network design for semantic segmentation. We employ a DeepLabV3 architecture \cite{chen2017deeplab}, with a VGG16 \cite{simonyan2015deep} feature extractor. Following the up-sampling decoder, we use a one-layer linear classifier. The network is trained on the source domain for $90,000$ iterations using the Adam optimizer. We use a learning rate of $1e-4$, decay of $1e-6$, $\epsilon=1e-6$, and batch size of $16$ images. When learning the GMM, we choose the high confidence parameter $\rho=.97$. The main results are reported for $t=3$ components for each semantic class. For adaptation, we initialize the network with source weights. We train for $35,000$ iterations, with a batch size of $32$ images. We again use the Adam optimizer, with $5e-5$ learning rate, decay of $1e-6$ and $\epsilon=1e-1$. The regularizer for the SWD loss, $\lambda$, is set to $.5$. In the batch selection process, the target image slices are selected at random, which can lead to scenarios where certain classes are missing, and batch distribution being significantly different from the target distribution. This can damage the adaptation process, leading to different semantic classes being merged together via the SWD loss. While a larger batch size can lead to the batch distribution becoming arbitrarily close to the target distribution, this is unfeasible when running on a single GPU. Thus, we address this issue by using the actual batch label distribution when selecting samples from the GMMs. The hardware we use is a Nvidia Titan Xp GPU. 






\subsection{Quantitative and Qualitative Results}

\begin{figure*}[!htb]
    \centering
    \includegraphics[width=.8\textwidth]{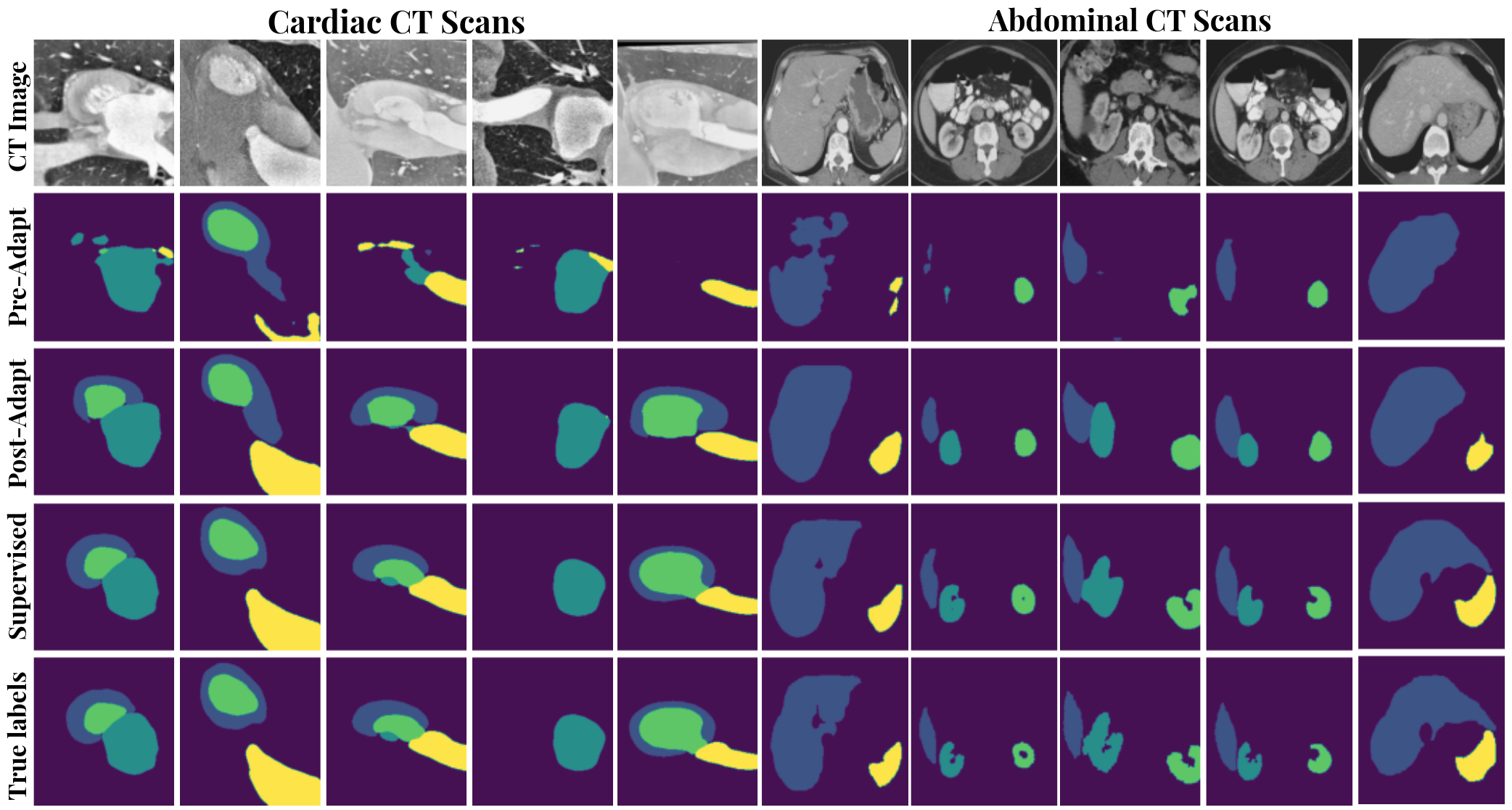}
    \caption{Segmentation maps of CT samples from the two datasets. The first five columns correspond to cardiac images, and last five correspond to abdominal images. From top to bottom: gray-scale CT images, source-only predictions, post-adaptation predictions, supervised predictions on the CT data, ground truth. }
    \label{figure:seg-maps}
\end{figure*}

We present our main quantitative results in Tables \ref{table:results-cardiac} through \ref{table:results-abdomen-ct-mri}. For methods which do not report cardiac and abdominal benchmarks, we report the performance from \cite{chen2020unsupervised}. The performance is upper bounded by the \textit{Supervised} benchmark, which corresponds to having full access to the target data for training a segmentation model. The \textit{Source-Only} benchmark corresponds to the model trained on the source domain directly applied to target data, which leads to poorest performance. On the cardiac dataset we observe that our method is able to outperform the other algorithms in terms of Dice score, with $81.3$ average performance on the $MRI\rightarrow CT$ task. We also achieve best performance on the $CT \rightarrow MRI$ task, with $72.3$ average Dice score. We restate that this is despite the fact that our method does not perform joint adaptation, losing access to the source domain after training. The GAN based methods offer competitive performance, as our approach achieves best class-wise performance only on the \textit{AA} class. We also compare positively to the other source free methods based on entropy minimization, showing that optimal-transport based adaptation has the potential of achieving better domain alignment. We further investigate GenAdapt \cite{Kundu_2021_ICCV}, a recent source-free algorithm for street semantic segmentation. The performance of this approach trails the other methods, confirming that the difference in data distribution between street images and medical images makes street semantic segmentation algorithms to not be directly transferable to the medical field. In terms of ASSD score, the best results corresponds to a GAN based method, however our method still offers competitive performance.

Our results on the abdominal organ dataset follow a similar trend, however on this secondary dataset the best performance is obtained by one of the GAN based methods. As mentioned before, our algorithm does not benefit from access to the source data during adaptation, thus the joint adaptation methods act as an upper-bound to our approach. Still, on the $MRI \rightarrow CT$ task we are able to obtain best performance on class \textit{Liver}, and close to SOTA performance on classes \textit{Left Kidney} and \textit{Spleen}. On the $CT \rightarrow MRI$ task we again observe competitive performance against the other methods and best performance on the \textit{Left Kidney} class, however deteriorated performance on class \textit{Spleen}. This shows that while overall performance is improved, due to the unsupervised nature of the adaptation process, there may be situations where the increase in accuracy is not similar across all classes.


\begin{table}[ht!]
  \begin{adjustbox}{center}
    \scalebox{1}{
    \setlength\tabcolsep{1.5pt} 
        \begin{tabular} { |c|cccc|c|cccc|c| }
            \hline
            &\multicolumn{5}{c|}{Dice} &\multicolumn{5}{c|}{ASSD} \\ \hline
            Method &AA &LAC &LVC &MYO &Avg. &AA &LAC &LVC &MYO &Avg. \\ \hline
            Source-Only &28.4 &27.7 &4.0 &8.7 &17.2 &20.6 &16.2 &N/A &48.4 &N/A \\ \hline
            Supervised$^*$ &88.7 &89.3 &89.0 &88.7 &87.2 &2.6 &4.9 &2.2 &1.6 &3.6 \\ \hline
            GenAdapt$^*$ \cite{Kundu_2021_ICCV} &57 &51 &36 &31 &43.8 &N/A &N/A &N/A &N/A &N/A \\ \hline
            PnP-AdaNet \cite{dou2019pnp} &74.0 &68.9 &61.9 &50.8 &63.9 &12.8 &6.3 &17.4 &14.7 &12.8 \\ \hline
            SynSeg-Net \cite{Huo_2019} &71.6 &69.0 &51.6 &40.8 &58.2 &11.7 &7.8 &7.0 &9.2 &8.9 \\ \hline
            AdaOutput \cite{Tsai_adaptseg_2018} &65.2 &76.6 &54.4 &43.3 &59.9 &17.9 &5.5 &5.9 &8.9 &9.6 \\ \hline
            CycleGAN \cite{zhu2020unpaired} &73.8 &75.7 &52.3 &28.7 &57.6 &11.5 &13.6 &9.2 &8.8 &10.8 \\ \hline
            CyCADA \cite{hoffman2018cycada} &72.9 &77.0 &62.4 &45.3 &64.4 &9.6 &8.0 &9.6 &10.5 &9.4 \\ \hline
            SIFA \cite{chen2020unsupervised} &81.3 &79.5 &73.8 &61.6 &74.1 &7.9 &6.2 &5.5 &8.5 &7.0 \\ \hline
            ARL-GAN \cite{chen2020anatomy} &71.3 &80.6 &69.5 &\textbf{81.6} &75.7 &6.3 &5.9 &6.7 &6.5 &6.4 \\ \hline
            DSFN \cite{ijcai2020-455} &84.7 &76.9 &79.1 &62.4 &75.8 &N/A &N/A &N/A &N/A &N/A \\ \hline
            SASAN \cite{tomar2021self} &82.0 &76.0 &82.0 &72.0 &78.0 &\textbf{4.1} &8.3 &\textbf{3.5} &\textbf{3.3} &\textbf{4.9} \\ \hline
            DSAN \cite{han2021deep} &79.9 &\textbf{84.7} &\textbf{82.7} &66.5 &78.5 &7.7 &6.7 &3.8 &5.6 &5.9 \\ \hline
            AdaEnt$^*$ \cite{BatesonSFDA} &75.5 &71.2 &59.4 &56.4 &65.6 &8.5 &7.1 &8.4 &8.6 &8.2 \\ \hline
            AdaMI$^*$ \cite{bateson2022source} &83.1 &78.2 &74.5 &66.8 &75.7 &5.6 &\textbf{4.2} &5.7 &6.9 &5.6 \\ \hline
            \hline 
            \textbf{SFS$^*$} &\textbf{88.0} &83.7 &81.0 &72.5 &\textbf{81.3} &6.3 &7.2 &4.7 &6.1 &6.1 \\
            \hline
          \end{tabular}
    }
  \end{adjustbox}
  \caption{Segmentation performance comparison for the \textbf{Cardiac} MR $\rightarrow$ CT adaptation task. Starred methods perform source-free adaptation. Bolded cells show best performance.}
  \label{table:results-cardiac}
\end{table}

\begin{table}[ht!]
  \begin{adjustbox}{center}
    \scalebox{1}{
    \setlength\tabcolsep{1.5pt} 
        \begin{tabular} { |c|cccc|c|cccc|c| }
            \hline
            &\multicolumn{5}{c|}{Dice} &\multicolumn{5}{c|}{ASSD} \\ \hline
            Method &AA &LAC &LVC &MYO &Avg. &AA &LAC &LVC &MYO &Avg. \\ 
            \hline
            Source only[6] &5.4 &30.2 &24.6 &2.7 &15.7 &15.4 &16.8 &13.0 &10.8 &14.0 \\
            Supervised[6] &82.8 &80.5 &92.4 &78.8 &83.6 &3.6 &3.9 &2.1 &1.9 &2.9 \\
            \hline
            PnP-AdaNet \cite{dou2019pnp} &43.7 &47.0 &77.7 &48.6 &54.3 &11.4 &14.5 &4.5 &5.3 &8.9 \\
            CyCADA\cite{hoffman2018cycada} &60.5 &44.0 &77.6 &47.9 &57.5 &7.7 &13.9 &4.8 &5.2 &7.9 \\
            SynSeg-Net\cite{Huo_2019} &41.3 &57.5 &63.6 &36.5 &49.7 &8.6 &10.7 &5.4 &5.9 &7.6 \\
            AdaOutput\cite{Tsai_adaptseg_2018} &60.8 &39.8 &71.5 &35.5 &51.9 &5.7 &8.0 &4.6 &4.6 &5.7 \\
            CycleGAN\cite{zhu2020unpaired} &64.3 &30.7 &65.0 &43.0 &50.7 &5.8 &9.8 &6.0 &5.0 &6.6 \\
            SIFA \cite{chen2020unsupervised} &65.3 &62.3 &78.9 &47.3 &63.4 &7.3 &7.4 &3.8 &4.4 &5.7 \\
            SASAN \cite{tomar2021self} &54 &\textbf{73} &86 &\textbf{68} &70 &18.8 &9.4 &6.1 &3.9 &9.5 \\
            DSAN \cite{han2021deep} &\textbf{71.3} &66.2 &76.2 &52.1 &66.5 &4.4 &7.3 &5.5 &4.3 &5.4 \\
            \hline
            \hline 
            \textbf{SFS$^*$} &66.4 &69.0 &\textbf{89.0} &64.5 &\textbf{72.3} &\textbf{3.13} &\textbf{2.8} &\textbf{0.33} &\textbf{1.16} &\textbf{1.56}  \\
            \hline
          \end{tabular}
         
    }
  \end{adjustbox}
  \caption{Segmentation performance comparison for the \textbf{Cardiac} CT $\rightarrow$ MR adaptation task.}
  \label{table:results-cardiac-ct-mri}
\end{table}

\begin{table}[ht!]
  \begin{adjustbox}{center}
    \scalebox{1}{
    \setlength\tabcolsep{1.5pt} 
        \begin{tabular} { |c|cccc|c|cccc|c| }
            \hline
            &\multicolumn{5}{c|}{Dice} &\multicolumn{5}{c|}{ASSD} \\ \hline
            Method &L &RK &LK &S &Avg. &L &RK &LK &S &Avg. \\
            \hline
            Source-Only &73.1 &47.3 &57.3 &55.1 &58.2 &2.9 &5.6 &7.7 &7.4 &5.9  \\ \hline
            Supervised &94.2 &87.2 &88.9 &89.1 &89.8 &1.2 &1.2 &1.1 &1.7 &1.3 \\ \hline
            SynSeg-Net \cite{Huo_2019} &85.0 &82.1 &72.7 &81.0 &80.2 &2.2 &1.3 &2.1 &2.0 &1.9 \\ \hline
            AdaOutput \cite{Tsai_adaptseg_2018} &85.4 &79.7 &79.7 &81.7 &81.6 &1.7 &1.2 &1.8 &1.6 &1.6 \\ \hline
            CycleGAN \cite{zhu2020unpaired} &83.4 &79.3 &79.4 &77.3 &79.9 &1.8 &1.3 &\textbf{1.2} &1.9 &1.6 \\ \hline
            CyCADA \cite{hoffman2018cycada} &84.5 &78.6 &80.3 &76.9 &80.1 &2.6 &1.4 &1.3 &1.9 &1.8 \\ \hline
            SIFA \cite{chen2020unsupervised} &88.0 &\textbf{83.3} &\textbf{80.9} &\textbf{82.6} &\textbf{83.7} &\textbf{1.2} &\textbf{1.0} &1.5 &\textbf{1.6} &\textbf{1.3} \\ \hline
            \hline
            \textbf{SFS$^*$} &\textbf{88.3} &73.7 &80.7 &81.6 &81.1 &2.4 &4.1 &3.5 &2.7 &3.2 \\
            \hline
          \end{tabular}
    }
  \end{adjustbox}
  \caption{Segmentation performance comparison for the Abdominal MR $\rightarrow$ CT adaptation task. }
  \label{table:results-abdomen}
\end{table}

\begin{table}[ht!]
  \begin{adjustbox}{center}
    \scalebox{1}{
    \setlength\tabcolsep{1.5pt} 
        \begin{tabular} { |c|cccc|c|cccc|c| }
            \hline
            &\multicolumn{5}{c|}{Dice} &\multicolumn{5}{c|}{ASSD} \\ \hline
            Method &L &RK &LK &S &Avg. &L &RK &LK &S &Avg. \\
            \hline
            Source-Only &48.9 &50.9 &65.3 &65.7 &57.7 &4.5 &12.3 &6.8 &4.5 &7.0  \\ \hline
            Supervised &92.0 &91.1 &80.6 &85.7 &87.3 &1.3 &2.0 &1.5 &1.3 &1.5 \\ \hline
            SynSeg-Net \cite{Huo_2019} &87.2 &90.2 &76.6 &79.6 &83.4 &2.8 &\textbf{0.7} &4.8 &2.5 &2.7 \\ \hline
            AdaOutput \cite{Tsai_adaptseg_2018} &85.8 &89.7 &76.3 &\textbf{82.2} &83.5 &1.9 &1.4 &3.0 &1.8 &2.1 \\ \hline
            CycleGAN \cite{zhu2020unpaired} &\textbf{88.8} &87.3 &76.8 &79.4 &83.1 &2.0 &3.2 &1.9 &2.6 &2.4 \\ \hline
            CyCADA \cite{hoffman2018cycada} &88.7 &89.3 &78.1 &80.2 &84.1 &\textbf{1.5} &1.7 &\textbf{1.3} &\textbf{1.6} &\textbf{1.5} \\ \hline
            SIFA \cite{chen2020unsupervised} &88.5 &\textbf{90.0} &79.7 &81.3 &\textbf{84.9} &2.3 &0.9 &1.4 &2.4 &1.7 \\ \hline
            \hline
            \textbf{SFS$^*$} &86.3 &88.0 &\textbf{85.1} &74.9 &83.5  &4.5 &1.6 &2.2 &18.2 &6.6  \\
            \hline
          \end{tabular}
    }
  \end{adjustbox}
  \caption{Segmentation performance comparison for the Abdominal CT $\rightarrow$ MRI adaptation task.}
  \label{table:results-abdomen-ct-mri}
\end{table}

Our results indicate that our method can also be used to perform UDA is a sequential setting, where the target domain becomes available later after the source domain data. The setting is similar to the continual learning setting and offers the prospect of adopting our method in this setting~\cite{rostami2020generative,rostami2020using}, while relaxing  the criteria that all tasks should be labeled.

In addition to quantitative verifying our approach on the two datasets, we also provide a visualization of the benefits of the adaptation process. In Figure \ref{figure:seg-maps} we display target images for the cardiac and abdominal datasets, and semantic maps for true labels, supervised predictions, adaptation predictions and source-only predictions. First, we note that the supervised performance is able to reach semantic map quality similar to the true labels, however incorrect labels are present with regards to the image borders, e.g. first three abdominal images. We believe this behavior stems from the resolution and architecture of the network. The \text{Pre-Adapt} semantic maps correspond to source-only performance, before model adaptation. These results present large inconsistencies with regards to the true label maps, and show the prediction deterioration due to domain-shift. The \textit{Post-Adapt} images clearly show improvement over source-only performance, and we observe the semantic maps become much closer to supervised performance. This process is however not perfect, as we notice difficulty in correctly determining some image borders, for example in the second cardiac image and in the first three abdominal images. This supports our quantitative evaluation, where we observed high Dice scores, however GAN based methods showed improved performance with respect to the ASSD metric.

\subsection{Ablation Studies and Analysis}

\begin{figure*}[!htb]
    \centering
    \subfloat[GMM samples]{
        \includegraphics[width=.23\textwidth]{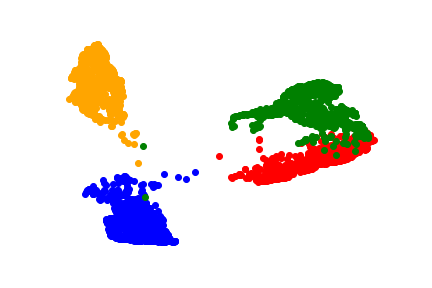}
        \label{figure:mmwhs-latent-features1}
    }
    \subfloat[Pre-adaptation]{
        \includegraphics[width=.23\textwidth]{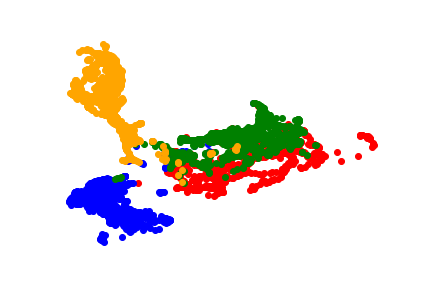}
        \label{figure:mmwhs-latent-features2}
    }
    \subfloat[Post-adaptation]{
        \includegraphics[width=.23\textwidth]{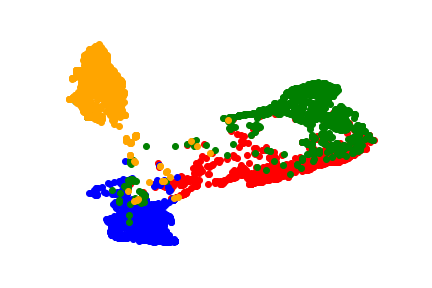}
        \label{figure:mmwhs-latent-features3}
    }
    \caption{Indirect distribution matching in the embedding space: (a) GMM samples approximating the MMWHS MR latent distribution, (b) CT latent embedding prior to adaptation (c) CT latent embedding post domain alignment. Colors correspond to: \textcolor{orange}{AA}, \textcolor{blue}{LAC}, \textcolor{green}{LVC}, \textcolor{red}{MYO}.}
    \label{figure:mmwhs-latent-features}
\end{figure*}

\begin{figure*}[!htb]
    \centering
    \subfloat[GMM samples]{
        \includegraphics[width=.23\textwidth]{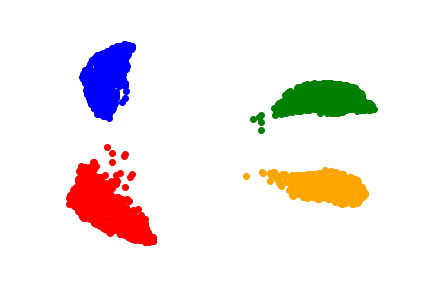}
        \label{figure:abdomen-latent-features1}
    }
    \subfloat[Pre-adaptation]{
        \includegraphics[width=.23\textwidth]{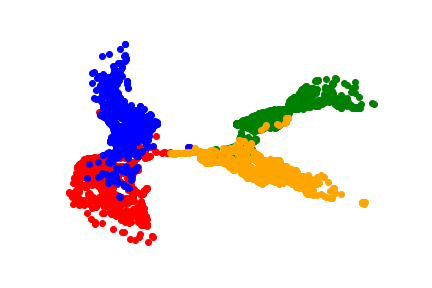}
        \label{figure:abdomen-latent-features2}
    }
    \subfloat[Post-adaptation]{
        \includegraphics[width=.23\textwidth]{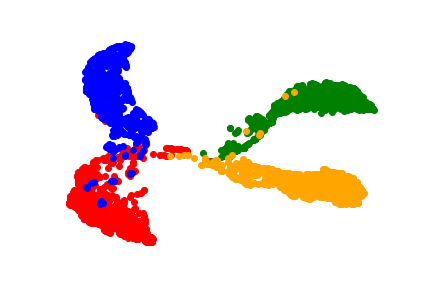}
        \label{figure:abdomen-latent-features3}
    }
    \caption{Indirect distribution matching in the embedding space: (a) GMM samples approximating the CHAOS MR latent distribution, (b) Multi-Atlas CT embedding prior to adaptation (c) Multi-Atlas CT embedding post adaptation. Colors correspond to: \textcolor{red}{liver}, \textcolor{blue}{right kidney}, \textcolor{green}{left kidney}, \textcolor{orange}{spleen}.} 
    \label{figure:abdomen-latent-features}
\end{figure*}

We analyze the impact of the adaptation process on the internal feature representations of our model. For this purpose, we employ the UMAP \cite{mcinnes2020umap} visualization tool to project the high dimensional feature representations to two dimensions. 

First, we investigate the latent distribution shift with respect to the source approximations for both the cardiac and abdominal datasets. Figures \ref{figure:mmwhs-latent-features}(a) and \ref{figure:abdomen-latent-features}(a) display the GMM embedding distributions approximating the source domain. The GMM distribution approximate the latent source feature space, and we observe that after source training this feature space indeed clusters into $K$ clusters. Figures \ref{figure:mmwhs-latent-features}(b) and \ref{figure:abdomen-latent-features}(b) correspond to the target embeddings before adaptation. We note that prior to adaptation, the target embeddings suffer from significant class overlap. Additionally, we expect that following the adaptation process, the target latent features will become more similar to the source latent features, in order to guarantee classifier generalization. We observe this behavior in figures \ref{figure:mmwhs-latent-features}(c) and \ref{figure:abdomen-latent-features}(c). For both datasets, the target embeddings become more similar to the GMM distribution, providing a visual confirmation that our distributional distance minimization process is able to reduce the source/target domain gap. 

\begin{figure}[!htb]
    \centering
    \subfloat[$\rho=0$]{
        \includegraphics[width=.2\textwidth]{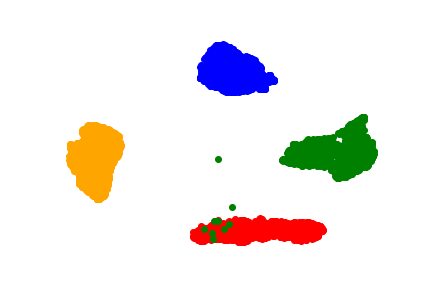}
        \label{figure:gaussians-0}
    }
    \subfloat[$\rho=0.8$]{
        \includegraphics[width=.2\textwidth]{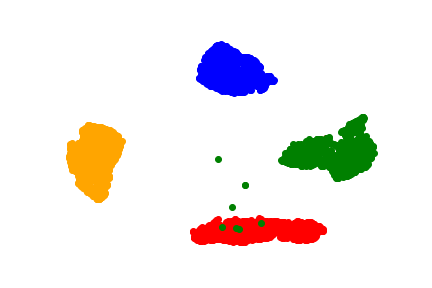}
        \label{figure:gaussians-8}
    }
    \subfloat[$\rho=.97$]{
        \includegraphics[width=.2\textwidth]{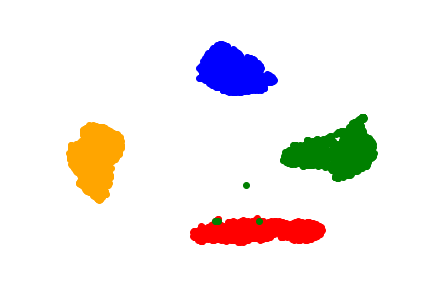}
        \label{figure:gaussians-97}
    }
    \caption{Learnt Gaussian embeddings on the cardiac dataset for different $\rho$.}
    \label{figure:mmwhs-gaussian-embeddings}
\end{figure}

We additionally investigate the impact of the $\rho$ parameter on the adaptation process. This parameter controls the GMM learning process by eliminating from it all source embeddings on which the classifier does not offer high confidence. Such samples are close to the classifier's decision boundary, and thus correspond to outlier class elements. Introducing these in the GMM learning process can have the negative effect of reducing class separability for samples from different Gaussians. In the adaptation process, this can in turn lead to more samples becoming closer to the decision function, and thus increasing the chance they would be miss-labeled. We verify this intuition by analyzing the GMM embeddings for three values of $\rho$ in Figure \ref{figure:mmwhs-gaussian-embeddings}. While the clusters are fairly well separated even for $\rho=0$, we there exists overlap between two of the classes. This overlap further decreases as $\rho$ is increased, such that for $\rho=.97$ it is greatly diminished. In our implementation, we observed little benefit once $\rho$ became larger than $.95$. 

\begin{figure}[!htb]
    \centering
    \subfloat[$\rho=0$]{
        \includegraphics[width=.35\textwidth]{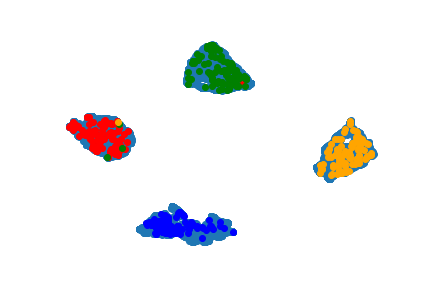}
        \label{figure:gaussians-0-over-source}
    }
    \subfloat[$\rho=.97$]{
        \includegraphics[width=.35\textwidth]{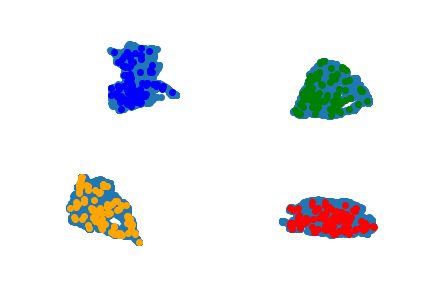}
        \label{figure:gaussians-97-over-source}
    }
    \caption{Learnt GMM embeddings plotted over source embeddings. Colored points denote the GMMs \& dark blue points denote the source samples.}
    \label{figure:mmwhs-gaussian-embeddings-over-source}
\end{figure}

We investigate the qualitative approximation of the source distributions via the intermediate distribution. While in Figures \ref{figure:mmwhs-latent-features}(a) and \ref{figure:abdomen-latent-features}(b) we observed that the learnt GMM distribution clusters over the semantic classes, this does not directly guarantee that the GMM offer a good approximation of the source embedding distribution. We verify this fact in Figure \ref{figure:mmwhs-gaussian-embeddings-over-source}, where we display the Gaussian embeddings for high-confidence thresholds $\rho=0$ and $\rho=.97$ along with source samples. When $\rho=0$, we use all available samples for learning the GMM, and when $\rho=.97$, only the high confidence samples are used. We observe the GMM distribution offers a good visual approximation of the source distribution in both cases. However, similar to Figure \ref{figure:mmwhs-gaussian-embeddings} , we notice that if $\rho$ is set to $0$, there will be GMM samples clustering under the wrong class, e.g. green and orange samples in the red cluster. This enforces the idea that including points near the decision boundary in the GMM learning process will in turn lead to more GMM samples close to the decision boundary, which may produce miss-classifications in the adaptation pipeline. We do not observe such behavior when the high confidence threshold is increased, hence the choice of $\rho=.97$.

\begin{figure}[!htb]
    \centering
    \includegraphics[width=.4\textwidth]{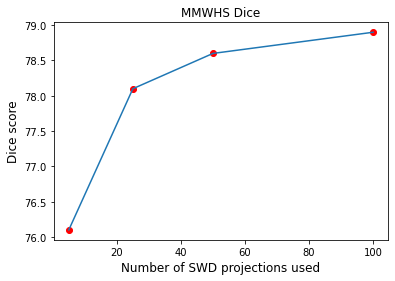}
    \caption{Dice scores on the cardiac task when $5, 25, 50$ and $100$ projections being used when computing the SWD.}
    \label{figure:w2-approximation}
\end{figure}

As presented in Eq. \ref{eq:swd}, the SWD is an approximation of optimal transport which relies on repeatedly projecting the two distributions into one dimension, and then computing the WD against these projections. The quality of the approximation of the WD by SWD is however tied to the number of projections used during this computation, with larger numbers of projections leading to larger similarity between the two metrics. In our work, we use $V=100$ projections, and choose this number by investigating several values of $V$. We report results in Figure \ref{figure:w2-approximation}, where we compute Dice scores on the MMWHS dataset for different numbers of projections. we observe if the number of projections is small, e.g. less than $10$, there is a significant performance drop in the reported result. However when $V$ increases, the difference in Dice performance become minimal. In our work, we choose $V=100$ when computing the SWD, however using a larger value of $V$ may further increase performance.

We further investigate the adaptation process in terms of class-wise pixel shift. In Tables \ref{table:mmwhs-percentage-migrate} and \ref{table:abdomen-percentage-migrate} for each row/column pair we report the percentage of pixels that change categories, the percentage out of these which change incorrectly and finally the percentage which change correctly. We note that when more than $1\%$ of the pixels change, the adaptation process always leads to a decrease in incorrectly labeled pixels. We notice that this happens more for the cardiac dataset compared to the abdominal dataset, which similar to Figures \ref{figure:mmwhs-latent-features},\ref{figure:abdomen-latent-features}, shows the cardiac datasets has less class separability than the abdominal dataset. We also note that for both datasets, a large number of samples that were labeled incorrectly before the adaptation correspond to the \textit{Ignore} class.

\begin{table}[ht!]
  \begin{adjustbox}{center}
    \scalebox{.75}{\small
    \setlength\tabcolsep{1.5pt} 
        \begin{tabular} {|c| *{5}{|ccc|}}
            \hline
            &\multicolumn{3}{|c||}{Ignore} &\multicolumn{3}{|c||}{MYO}
            &\multicolumn{3}{|c||}{LAC} &\multicolumn{3}{|c||}{LVC}
            &\multicolumn{3}{|c|}{AA} \\ \hline
            Ignore &\textbf{97.3} &\textbf{99.3} &\textbf{99.3} &\textbf{1.5} &\textbf{20.3} &\textbf{70.0} &0.2 &80.2 &14.8 &0.9 &6.2 &76.1 &0.2 &43.8 &51.7 \\ \hline
MYO &\textbf{13.2} &\textbf{10.4} &\textbf{89.5} &\textbf{81.6} &\textbf{72.2} &\textbf{72.2} &0.1 &52.7 &0.4 &\textbf{5.2} &\textbf{44.6} &\textbf{54.1} &0.0 &0.0 &0.0 \\ \hline
LAC &\textbf{15.1} &\textbf{45.4} &\textbf{46.3} &\textbf{2.5} &\textbf{2.6} &\textbf{79.7} &\textbf{76.1} &\textbf{88.4} &\textbf{88.4} &\textbf{5.9} &\textbf{7.4} &\textbf{87.4} &0.4 &5.8 &77.0 \\ \hline
LVC &0.6 &67.7 &2.3 &\textbf{16.5} &\textbf{33.4} &\textbf{66.3} &0.2 &83.8 &13.0 &\textbf{82.7} &\textbf{92.4} &\textbf{92.4} &0.0 &93.3 &0.0 \\ \hline
AA &\textbf{18.5} &\textbf{7.8} &\textbf{90.9} &0.0 &0.0 &43.7 &\textbf{1.3} &\textbf{5.7} &\textbf{6.2} &0.1 &0.0 &12.9 &\textbf{80.1} &\textbf{91.2} &\textbf{91.2} \\ \hline
        \end{tabular}
    }
  \end{adjustbox}
  \caption{Percentage of shift in pixel labels during adaptation for the cardiac dataset. A cell $(i,j)$ in the table has three values. The first value represents the percentage of pixels labeled $i$ that are labeled $j$ after adaptation. The second value represents the percentage of switching pixels whose true label is $i$ - lower is better. The third value represents the percentage of switching pixels whose true label is $j$ - higher is better. Bolded cells denote label shift where more than $1\%$ of pixels migrate from $i$ to $j$.}
  \label{table:mmwhs-percentage-migrate}
\end{table}

\begin{table}[ht!]
  \begin{adjustbox}{center}
    \scalebox{.72}{\small
    \setlength\tabcolsep{1.5pt} 
        \begin{tabular} {|c| *{5}{|ccc|}}
            \hline
            &\multicolumn{3}{|c||}{Ignore} &\multicolumn{3}{|c||}{Liver}
            &\multicolumn{3}{|c||}{R. Kidney} &\multicolumn{3}{|c||}{L. Kidney}
            &\multicolumn{3}{|c|}{Spleen} \\ \hline
            Ignore &\textbf{94.6} &\textbf{98.4} &\textbf{98.4} &\textbf{3.0} &\textbf{18.0} &\textbf{81.6} &0.7 &23.5 &74.3 &0.7 &34.9 &62.6 &\textbf{1.0} &\textbf{19.3} &\textbf{80.5} \\ \hline
Liver &\textbf{6.6} &\textbf{38.1} &\textbf{60.8} &\textbf{92.6} &\textbf{91.3} &\textbf{91.3} &0.8 &10.4 &55.1 &0.0 &0.0 &0.0 &0.0 &39.0 &10.2 \\ \hline
RKidney &\textbf{5.0} &\textbf{13.1} &\textbf{86.9} &0.2 &0.0 &76.9 &\textbf{94.8} &\textbf{94.7} &\textbf{94.7} &0.0 &0.0 &0.0 &0.0 &0.0 &0.0 \\ \hline
LKidney &\textbf{2.2} &\textbf{24.2} &\textbf{75.0} &0.1 &0.0 &0.0 &0.0 &23.7 &0.0 &\textbf{97.5} &\textbf{87.8} &\textbf{87.8} &0.2 &0.0 &7.2 \\ \hline
Spleen &\textbf{23.1} &\textbf{20.8} &\textbf{79.2} &0.1 &20.2 &0.0 &0.2 &75.0 &0.0 &0.0 &69.4 &0.0 &\textbf{76.6} &\textbf{78.7} &\textbf{78.7} \\ \hline
        \end{tabular}
    }
  \end{adjustbox}
  \caption{Percentage of shift in pixel labels during adaptation for the abdominal organ dataset. The same methodology as in Table \ref{table:mmwhs-percentage-migrate} is used.}
  \label{table:abdomen-percentage-migrate}
\end{table}

Next, we analyze the impact of the parameter $t$ on the adaptation performance. In our GMM learning process, $t$ controls how many components are employed for each semantic class. While the minimum value of $t$ is $1$, a larger number of components should lead to a better approximation of the underlying distribution. We verify this idea in Table \ref{table:results-t-components}. We notice that as we increase $t$, the adaptation performance improves when keeping the number of training iterations fixed. This implies that the learnt GMM distribution captures more information from the source domain, which benefits the SWD distribution alignment process. In our main results in Tables \ref{table:results-cardiac} and \ref{table:results-abdomen} we utilize a $t$ value of $3$, as we see larger values of $t$ provide reduced benefit gains. 

\begin{table}[ht!]
  \begin{adjustbox}{center}
    \scalebox{.9}{
    \small
    \setlength\tabcolsep{1.5pt} 
        \begin{tabular} { |c|cccc|c|cccc|c| }
            \hline
            &\multicolumn{5}{c|}{Dice} &\multicolumn{5}{c|}{Average Symmetric Surface Distance} \\ \hline
            Method &AA &LAC &LVC &MYO &Average &AA &LAC &LVC &MYO &Average \\ \hline
            \textbf{1-SFS} &86.2 &83.5 &75.4 &70.9 &79.0 &11.1 &5.0 &10.8 &3.6 &9.8 \\
            \hline
            \textbf{3-SFS} &88.0 &83.7 &81.0 &72.5 &81.3 &6.3 &7.2 &4.7 &6.1 &6.1 \\
            \hline
            \textbf{5-SFS} &88.0 &83.8 &81.9 &73.3 &81.7 &6.2 &7.4 &4.8 &5.7 &6.0 \\
            \hline
            \textbf{7-SFS} &86.8 &84.8 &82.0 &73.5 &81.8 &4.8 &7.2 &4.4 &5.6 &5.9 \\
            \hline
          \end{tabular}
    }
  \end{adjustbox}
  \caption{Segmentation performance comparison for the Cardiac MR $\rightarrow$ CT adaptation task. $t$-SFS represents results for $t$ components per class. }
  \label{table:results-t-components}
\end{table}

Finally, we compare our distributional alignment metric, the SWD, to other popular distributional alignment approaches. We consider the KL-divergence and MMD as additional comparison metrics. We report results of using these two metrics for adaptation on the cardiac dataset in Table \ref{table:results-different-distances}. We note using MMD does not benefit our framework. Using the KL divergence metric provides a positive outcome, and we obtain a result close to the one via SWD, however with slightly deteriorated performance. This empirically shows that SWD is more robust than some other distributional distance metrics.

\begin{table}[ht!]
  \begin{adjustbox}{center}
    \scalebox{.9}{
    \small
    \setlength\tabcolsep{1.5pt} 
        \begin{tabular} { |c|cccc|c|cccc|c| }
            \hline
            &\multicolumn{5}{c|}{Dice} &\multicolumn{5}{c|}{Average Symmetric Surface Distance} \\ \hline
            Method &AA &LAC &LVC &MYO &Average &AA &LAC &LVC &MYO &Average \\ \hline
            \textbf{MDD} &24.4 &73.3 &19.1 &68.1 &46.2 &21.1 &4.8 &33.3 &20.3 &19.9  \\
            \textbf{KL} &87.9 &87.3 &74.7 &62.7 &78.1 &8.4 &6.6 &11.7 &9.6 &9.1 \\
            \hline
            \textbf{SFS} &88.0 &83.7 &81.0 &72.5 &81.3 &6.3 &7.2 &4.7 &6.1 &6.1 \\
            \hline
          \end{tabular}
    }
  \end{adjustbox}
  \caption{Segmentation performance on the Cardiac MR $\rightarrow$ CT adaptation task for different distributional distances. }
  \label{table:results-different-distances}
\end{table}

\section{Conclusion}

We have presented a source-free UDA algorithm for the segmentation of medical images. We proved our algorithm minimizes an upper bound on the target error, and showed that it empirically compares favorably to source-free and joint UDA medical imaging works on a cardiac and internal organ dataset. We validated the adaptation process by further analyzing the changes in distribution shift performed by various components of our approach, and showed that the empirical observations align with the intuition behind the algorithm.

 {
     
    \small
    \bibliographystyle{plain}
    \bibliography{ref}
}

\end{document}